\pdfoutput=1

\documentclass[11pt]{article}

\usepackage{ACL2023}

\usepackage{times}
\usepackage{latexsym}
\usepackage{diagbox} 

\usepackage[T1]{fontenc}
\usepackage[utf8]{inputenc}
\usepackage{hyperref} 

\usepackage{microtype}
\usepackage{colortbl} 

\usepackage{inconsolata}
\usepackage{multirow}
\usepackage{graphicx}
\usepackage{amsmath}

\usepackage{amssymb} 
\usepackage{xcolor} 
\usepackage{booktabs} 
\usepackage{arydshln} 
\usepackage{subcaption}
\usepackage{needspace}
\usepackage{hyperref}

\usepackage[most]{tcolorbox} 

\usepackage{listings}
\definecolor{softgreen}{rgb}{0.0, 0.5, 0.0} 

\usepackage{titlesec}

\usepackage{enumitem}

\title{Learning or Self-aligning? Rethinking Instruction Fine-tuning}

\author{Mengjie Ren${}^{1,3}$\thanks{~ Work was partially done during Ren's internship at Meituan.}, Boxi Cao${}^{1,3}$, Hongyu Lin${}^{1}$\thanks{~ Corresponding Author}, Cao Liu${}^{4}$, Xianpei Han${}^{1,2,5}$\\
\textbf{Ke Zeng}${}^{4}$, \textbf{Guanglu Wan}${}^{4}$, \textbf{Xunliang Cai}${}^{4}$, \textbf{Le Sun}${}^{1,2,5}$\\
${}^{1}$Chinese Information Processing Laboratory ~ ${}^{2}$State Key Laboratory of Computer Science \\
Institute of Software, Chinese Academy of Sciences\\
${}^{3}$University of Chinese Academy of Sciences\\
${}^{4}$Meituan\\
${}^{5}$Key Laboratory of System Software, Chinese Academy of Sciences \\
{\tt \{renmengjie2021,boxi2020,hongyu,xianpei,sunle\}@iscas.ac.cn} \\
{\tt \{liucao,zengke02,wanguanglu,caixunliang\}@meituan.com }}

\begin{document}
\maketitle

\begin{abstract}

Instruction Fine-tuning~(IFT) is a crucial phase in building large language models~(LLMs). 
Previous works mainly focus on the IFT's role in the transfer of behavioral norms and the learning of additional world knowledge. 
However, the understanding of the underlying mechanisms of IFT remains significantly limited. 
In this paper, we design a knowledge intervention framework to decouple the potential underlying factors of IFT, thereby enabling individual analysis of different factors.
Surprisingly, our experiments reveal that attempting to learn additional world knowledge through IFT often struggles to yield positive impacts and can even lead to markedly negative effects. 
Further, we discover that maintaining internal knowledge consistency before and after IFT is a critical factor for achieving successful IFT. Our findings reveal the underlying mechanisms of IFT and provide robust support for some very recent and potential future works.
We release our experimental dataset and codes to facilitate future work~\footnote{\href{https://github.com/renmengjie7/Self-Aligning}{https://github.com/renmengjie7/Self-Aligning}}.
\end{abstract}

\section{Introduction}
\label{sec:Introduction}

The advent and evolution of large language models~(LLMs) have marked a significant milestone in natural language processing~(NLP)~\cite{brown2020languagea, touvron2023llama, touvron2023llamab}.
As one of the core steps in the construction of LLMs, instruction fine-tuning (IFT) employs supervised instruction-response pairs to fine-tune LLMs, thereby facilitating the transformation of LLM from a continuous writing model to a question-answering agent~\cite{chung2024scaling, iyer2023optiml, jang2023exploring}.

\begin{figure}[!tp]
\centering
\includegraphics[width=\columnwidth]{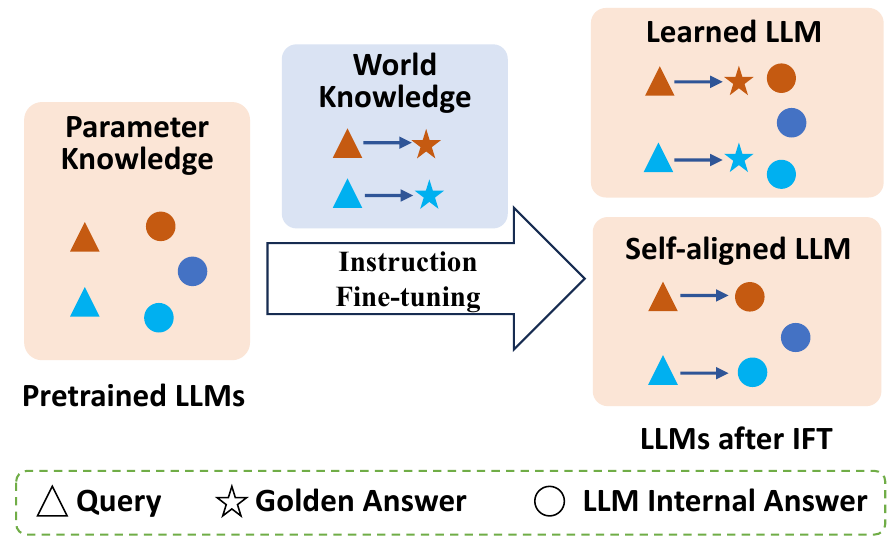}
\caption{
Two potential mechanisms for instruction fine-tuning. 1) Learning, which injects world knowledge in IFT data into LLMs; 2) Self-aligning, which aligns queries with knowledge already in LLMs with similar behavioral norms.
Elements with the same color are related.
}
\label{fig:intro}
\end{figure}

Despite the crucial role of IFT in the construction of LLMs, there is a significant lack of in-depth research on the mechanisms by which IFT operates. 
In traditional machine learning, supervised learning aims to fit models to specific tasks and data distributions~\cite{goodfellow2016deep,MF}, whereas the impact of IFT on LLMs is markedly different. 
As shown in Figure~\ref{fig:intro}, on one hand, one of the most apparent effects of IFT is its ability to align the output of LLMs more closely with the latent behavioral norms contained within the IFT data, thereby enabling more effective parameter knowledge expression~\cite{zhou2024lima, chen2023maybe}. 
On the other hand, many existing studies~\cite{li2023chatdoctor, cui2023chatlaw, chen2023soulchat} aim to facilitate domain-specific adaptation of LLMs through IFT, by injecting the world knowledge contained in IFT data into LLMs.

Unfortunately, both the transfer of behavioral norms and the enhancement of domain knowledge are closely coupled with the corpus applied in IFT, rendering the analysis of IFT's true effects exceedingly challenging. 
Due to the interconnection of these two effects, it is challenging to discern whether the benefits derived are due to the promotion of more effective expression of parameter knowledge or the injection of additional world knowledge.
The coupling between the above two factors, along with a lack of in-depth analysis of the IFT's mechanism, hampers our comprehension of the effectiveness of IFT. 
This limitation hinders our ability to develop robust strategies for IFT data construction, model training, and model evaluation due to insufficient theoretical support. 
Therefore, a thorough and comprehensive analysis of the underlying core factors that drive the effectiveness of IFT is crucial for achieving more effective IFT.

To this end, this paper designs a knowledge intervention framework for analyzing the underlying mechanisms of IFT. 
The main idea of our framework is to control the consistency between the knowledge in IFT data and the existing parameter knowledge of LLM, in order to decouple the injection of domain knowledge from the transfer of behavioral norms during IFT. 
This allows for a separate analysis of the roles of these two crucial factors. 
Specifically, we first employ in-context learning~(ICL)~\cite{dong2022survey, brown2020languagea} to probe the internal parameter knowledge of LLMs. 
Building on this, we intervene in the composition of existing parameter knowledge and the newly introduced world knowledge within IFT data and then observe the differences in model behavior after IFT using different intervention groups. 
Based on the framework, we conduct an in-depth analysis to answer the following two critical research questions~(RQ):

\begin{itemize}

\item \hypertarget{RQ1}{\textbf{RQ1: }\textit{How does the world knowledge within IFT data affect LLMs?}}
\item \hypertarget{RQ2}{\textbf{RQ2: }\textit{What is the underlying cause of the above impact?}}

\end{itemize}

For RQ1, we initially discover that significant discrepancies between the world knowledge contained in IFT data and the existing parameter knowledge within LLM can substantially undermine the model's abilities. 
Performance derived from a set of IFT data that contains incorrect world knowledge but aligns with the model's parameter knowledge is significantly better than that from a set containing correct world knowledge but inconsistent with the model's internal parameter knowledge. 
To dive into this phenomenon, we explicitly supply LLMs with the world knowledge necessary for answering the instruction, integrated into the context. 
This strategy allows the model to focus on transforming the output behavioral norms instead of jointly learning the inconsistent world knowledge. 
We discover that the detrimental effects caused by the inconsistency between parameter knowledge and world knowledge can be significantly mitigated by explicitly providing such self-contained IFT data points. 
These two findings indicate that attempting to introduce world knowledge through IFT that is inconsistent with the model's parameter knowledge can severely undermine the model, which suggests that the injection of world knowledge does not lie at the center of a successful IFT.

For RQ2, we further analyze the model performance under varying degrees of consistency between the knowledge contained in IFT data and the parameter knowledge in the original LLM. 
We find that while the consistency between the two has a significant impact on model performance, a higher degree of consistency does not necessarily correlate with better model performance. 
However, our further research reveals a strong correlation between the model's ultimate performance after IFT and the consistency of the model's internal knowledge before and after IFT. 
That is to say, for the model after IFT, if its responses are more consistent with the responses produced by the original model from in-context learning probing, then the performance of the model after IFT is also better. 
The validity of the finding is independent of whether the test data belongs to the same domain as the training data, and is also unrelated to the original performance of the base LLM. 
This implies that the phenomenon is solely influenced by the consistency of the knowledge before and after IFT.
Furthermore, we discover that using IFT data that is either too consistent or too inconsistent with the original parameter knowledge can lead to a divergence in the model's internal knowledge before and after IFT, thereby resulting in a decline in performance.

Our experiments reveal the fundamental role of IFT in the construction of LLMs.
Essentially, IFT is not a supervised, domain-specific \textbf{learning} process, but a process of \textbf{self-aligning} instruction with existing internal knowledge of LLMs that can be obtained through few-shot in-context learning probing.
Our findings not only provide robust theoretical support for the very recently emerged research on self-alignment~\cite{sun2024principle} and super-alignment~\cite{burns2023weaktostrong}, etc., but also shed light on the future direction of data construction, model training, and model evaluation for IFT.

\section{Related Work}
\label{sec:Related work}
By observing output token distribution shift of models before and after IFT, \citet{lin2023unlocking} found that most shifts occur with stylistic tokens, strongly supporting the superficial alignment hypothesis~\citep{zhou2024lima}, false promise~\citep{gudibande2023false} and related works on IFT data construction~\cite{chen2023maybe,shen2024rethinking} and proxy-guided decoding~\citep{liu2024tuning}.
While providing intuitive insights, they fall short of providing a comprehensive analysis of IFT's underlying mechanisms.

Meanwhile, recent efforts have focused on achieving automated alignment, such as self-instruction-tuning~\cite{sun2024principle,guo2024human}, self-rewarding~\cite{yuan2024selfrewarding} and super-alignment~\cite{burns2023weaktostrong}. 
Despite repeated validations of their effectiveness, there remains limited understanding of their success.

\section{Knowledge Intervention Framework}
\label{sec:framework}

During the process of IFT, the potential transfer of behavioral norms and the injection of world knowledge are coupled together. 
Consequently, prior research on IFT has struggled to distinguish the relative effects of these two. 
To further investigate the underlying mechanisms of IFT, this paper designs a knowledge intervention framework to decouple these two factors.
The main idea behind our framework is to control the association between knowledge in IFT data and the existing parameter knowledge in LLM, thereby managing the degree of potential world knowledge that would be injected during IFT.
Through this, we can decouple the injection of world knowledge and the transfer of behavioral norms, by observing the effects of IFT at varying degrees of world knowledge injection.

Specifically, we select four multi-choice datasets from different domains.
For each question in each dataset, we employ in-context learning to probe the internal parameter knowledge for each base LLM. 
Then, we construct multiple instruction datasets by adjusting the consistency of the knowledge within IFT data and model parameter knowledge for each base model. 
Finally, we analyze the underlying impact of different degrees of world knowledge injection on IFT by fine-tuning LLMs under different settings and examining their performance on the homogeneous, in-domain, and out-of-domain test sets. 
In the following, we will introduce our knowledge intervention framework in detail.

\subsection{Domain IFT Corpus Setup}
In order to facilitate more efficient knowledge consistency identification and IFT evaluation, we select all IFT corpus in the form of multiple choice from four domains: medicine, history, engineering, and jurisprudence.
For medicine, we craft a dataset with 20,000 training, 2,206 testing, and 10 development instances by filtering MedMCQA~\cite{pal2022medmcqa} for entries with explanations and one correct answer, and then applying random sampling.
For the other three domains, we procure the relevant items from Xiezhi Benchmark~\cite{gu2024xiezhi} and held 10 for development and 250 for test~\footnote{The availability of multiple-choice datasets tailored to specific domains is extremely limited. Consequently, from this benchmark, we select the three domains with the largest volume of data.}.

To make a comprehensive evaluation, for each domain's IFT, we construct three types of test sets: \textbf{1) homogeneous test set (HOMO)}, which is held out from the same multiple-choice dataset of the domain; \textbf{2) in-domain test set (ID)}, including data from MMLU~\cite{hendrycks2021measuring} that belong to the domain; \textbf{3) an out-of-domain test set (OOD)}, comprising instances in MMLU that are from distinct domains. 
By observing the accuracy performance differences across the three types of test sets, we aim to examine the impact of IFT on various scenarios.
Please refer to the Appendix~\ref{sec:appendix Data Construction} for more details about our data processing.

\subsection{Parameter Knowledge Probing via Few-shot In-context Learning}
Our knowledge intervention framework relies on effectively detecting the parameter knowledge of pre-trained LLMs.
To this end, this paper leverages few-shot in-context learning~\cite{dong2022survey,brown2020languagea}, which is a widely-used approach for probing the abilities and internal knowledge of pre-trained LLMs~\cite{zhang2023rtuning,wan2024mitigating}, to identify the parameter knowledge of our base LLMs. 
Specifically, we utilize in-context learning to probe the base model's response to each data item in domain multi-choice dataset and regard the response as the model's parameter knowledge for this question.

\subsection{Construction of Instruction Data}
Upon probing the internal knowledge of pre-trained LLMs, we build instances based on the consistency between world knowledge contained within the domain data and parameter knowledge, thereby constructing different IFT datasets. 
Specifically, for each domain and base model, we construct IFT data under three settings, including:

\begin{itemize}
    \item \textbf{Harmonious setting}, which consists of data where the embedded world knowledge is consistent with model parameter knowledge. This means the pre-trained LLM can answer correctly under in-context learning.
    In the learning process under this setting, the model only needs to transfer behavioral norms, without the need to acquire additional world knowledge due to the above consistency.
    \item \textbf{Incompatible setting}, which comprises instances where the pre-trained LLM cannot correctly answer under in-context learning.
    Due to the complete inconsistency, the model needs to learn both the behavioral norms and the world knowledge during its training phase.
    \item \textbf{Self-aligning setting}, which consists of data in which the queries are exactly the same as those in the incompatible set, but we modify the answers corresponding to each query to match the pre-trained LLM's internal knowledge.
    Therefore, under this setting, all responses are incorrect, and the model will not learn any additional world knowledge.
\end{itemize}
To ensure a fair comparison, we maintain the same size across the three groups of data.
Meanwhile, to prevent the potential collapse during model training due to the exclusive use of multiple-choice questions, we generate an explanation for the answer and incorporate an equal proportion of general instruction data sampled from alpaca-gpt4-en~\cite{peng2023instruction}, thereby ensuring a more stable and real IFT.

\subsection{Experiment Setup}

\paragraph{Base Model}
We use LLaMA-2-7B, LLaMA-2-13B, LLaMA-2-70B~\cite{touvron2023llama}, and Mistral-7B~\cite{jiang2023mistral} as the base models of our experiments.

\paragraph{Training Details}
We only calculate loss on outputs, setting epoch to 3, learning rate to $2e^{-5}$ and batch size to 256. 
For Mistral-7B, we set learning rate to $1e^{-5}$ \footnote{Training loss of Mistral-7B using learning rate $2e^{-5}$ does not converge and even spikes.}.
We use DeepSpeed ZeRO3~\cite{rasley2020deepspeed} for LLaMA-2-70B and FSDP~\cite{zhao2023pytorch} for the other three.
All experiments are implemented on Nvidia A100-80GB GPUs.

\section{Exp-\textup{I}: Does Learning Domain-specific World Knowledge Matter for IFT?}
\label{exp1}

\begin{table*}[!ht]
\centering
\resizebox{\textwidth}{!}{%
\begin{tabular}{lcccccccccccc}
\toprule

\multirow{2}{*}{\textbf{Eval}} 
& \multicolumn{3}{c}{\textbf{Medicine}} 
& \multicolumn{3}{c}{\textbf{History}} 
& \multicolumn{3}{c}{\textbf{Engineering}} 
& \multicolumn{3}{c}{\textbf{Jurisprudence}} \\

\cmidrule(lr){2-4}
\cmidrule(lr){5-7}
\cmidrule(lr){8-10}
\cmidrule(lr){11-13}

& \textbf{HAR} & \textbf{INC} & \textbf{SELF} 
& \textbf{HAR} & \textbf{INC} & \textbf{SELF} 
& \textbf{HAR} & \textbf{INC} & \textbf{SELF} 
& \textbf{HAR} & \textbf{INC} & \textbf{SELF} \\

\midrule 
\rowcolor{gray!20} \multicolumn{13}{c}{\textbf{LLaMA-2-7B}} \\ 
\midrule

HOMO & $\textbf{40.22}_{11.77\uparrow}$ & 28.45 & $\underline{37.00}_{\hspace{0.4em}8.55\uparrow}$ & $\textbf{38.80}_{\hspace{0.4em}9.20\uparrow}$ & 29.60 & $\underline{33.60}_{\hspace{0.4em}4.00\uparrow}$ & $\textbf{48.40}_{16.00\uparrow}$ & 32.40 & $\underline{32.80}_{\hspace{0.4em}0.40\uparrow}$ & $\textbf{37.60}_{\hspace{0.4em}3.60\uparrow}$ & \underline{34.00} & $33.20_{\hspace{0.4em}0.80\downarrow}$ \\ 

 ID & $\underline{39.82}_{\hspace{0.4em}2.56\uparrow}$ & 37.26 & $\textbf{41.46}_{\hspace{0.4em}4.20\uparrow}$ & $\textbf{54.30}_{23.22\uparrow}$ & 31.08 & $\underline{46.02}_{14.94\uparrow}$ & $\textbf{42.07}_{11.04\uparrow}$ & \underline{31.03} & $26.21_{\hspace{0.4em}4.82\downarrow}$ & $\textbf{38.86}_{\hspace{0.4em}3.16\uparrow}$ & 35.70 & $\underline{36.34}_{\hspace{0.4em}0.64\uparrow}$ \\ 

 OOD & $\underline{39.97}_{\hspace{0.4em}3.22\uparrow}$ & 36.75 & $\textbf{40.94}_{\hspace{0.4em}4.19\uparrow}$ & $\textbf{39.64}_{\hspace{0.4em}8.95\uparrow}$ & 30.69 & $\underline{37.22}_{\hspace{0.4em}6.53\uparrow}$ & $\textbf{40.38}_{12.12\uparrow}$ & 28.26 & $\underline{29.17}_{\hspace{0.4em}0.91\uparrow}$ & $\textbf{38.49}_{\hspace{0.4em}3.93\uparrow}$ & 34.56 & $\underline{34.88}_{\hspace{0.4em}0.32\uparrow}$ \\

\midrule 
\rowcolor{gray!20} \multicolumn{13}{c}{\textbf{LLaMA-2-13B}} \\ 
\midrule

HOMO & $\textbf{40.83}_{\hspace{0.4em}4.78\uparrow}$ & \underline{36.05} & $34.41_{\hspace{0.4em}1.64\downarrow}$ & $\textbf{48.40}_{16.00\uparrow}$ & 32.40 & $\underline{43.60}_{11.20\uparrow}$ & $\textbf{58.00}_{20.80\uparrow}$ & 37.20 & $\underline{55.20}_{18.00\uparrow}$ & $\textbf{44.00}_{11.60\uparrow}$ & 32.40 & $\underline{37.60}_{\hspace{0.4em}5.20\uparrow}$ \\ 

 ID & $\textbf{55.43}_{20.37\uparrow}$ & 35.06 & $\underline{52.13}_{17.07\uparrow}$ & $\textbf{68.28}_{22.15\uparrow}$ & 46.13 & $\underline{64.09}_{17.96\uparrow}$ & $\textbf{45.52}_{15.86\uparrow}$ & 29.66 & $\underline{40.00}_{10.34\uparrow}$ & $\textbf{54.77}_{16.22\uparrow}$ & 38.55 & $\underline{52.77}_{14.22\uparrow}$ \\ 

 OOD & $\textbf{54.21}_{18.44\uparrow}$ & 35.77 & $\underline{50.98}_{15.21\uparrow}$ & $\textbf{51.30}_{13.32\uparrow}$ & 37.98 & $\underline{49.06}_{11.08\uparrow}$ & $\textbf{52.15}_{16.21\uparrow}$ & 35.94 & $\underline{51.12}_{15.18\uparrow}$ & $\textbf{50.83}_{11.57\uparrow}$ & 39.26 & $\underline{48.27}_{\hspace{0.4em}9.01\uparrow}$ \\

\midrule 
\rowcolor{gray!20} \multicolumn{13}{c}{\textbf{LLaMA-2-70B}} \\ 
\midrule

HOMO & $\textbf{47.95}_{\hspace{0.4em}5.41\uparrow}$ & 42.54 & $\underline{46.03}_{\hspace{0.4em}3.49\uparrow}$ & $\textbf{59.20}_{17.20\uparrow}$ & 42.00 & $\underline{51.60}_{\hspace{0.4em}9.60\uparrow}$ & $\textbf{62.40}_{\hspace{0.4em}7.20\uparrow}$ & 55.20 & $\underline{57.60}_{\hspace{0.4em}2.40\uparrow}$ & $\textbf{55.20}_{\hspace{0.4em}7.60\uparrow}$ & 47.60 & $\underline{51.60}_{\hspace{0.4em}4.00\uparrow}$ \\ 

 ID & $\textbf{65.37}_{\hspace{0.4em}3.97\uparrow}$ & 61.40 & $\underline{63.11}_{\hspace{0.4em}1.71\uparrow}$ & $\textbf{82.37}_{11.08\uparrow}$ & 71.29 & $\underline{81.29}_{10.00\uparrow}$ & $\textbf{55.17}_{15.86\uparrow}$ & 39.31 & $\underline{54.48}_{15.17\uparrow}$ & $\textbf{67.69}_{\hspace{0.4em}5.48\uparrow}$ & 62.21 & $\underline{67.52}_{\hspace{0.4em}5.31\uparrow}$ \\ 

 OOD & $\textbf{65.34}_{\hspace{0.4em}4.99\uparrow}$ & 60.35 & $\underline{63.93}_{\hspace{0.4em}3.58\uparrow}$ & $\textbf{63.63}_{\hspace{0.4em}5.69\uparrow}$ & 57.94 & $\underline{63.54}_{\hspace{0.4em}5.60\uparrow}$ & $\textbf{65.62}_{\hspace{0.4em}6.41\uparrow}$ & 59.21 & $\underline{64.75}_{\hspace{0.4em}5.54\uparrow}$ & $\textbf{61.90}_{\hspace{0.4em}4.87\uparrow}$ & 57.03 & $\underline{61.45}_{\hspace{0.4em}4.42\uparrow}$ \\

\midrule 
\rowcolor{gray!20} \multicolumn{13}{c}{\textbf{Mistral-7B}} \\ 
\midrule

HOMO & $\textbf{49.80}_{15.12\uparrow}$ & 34.68 & $\underline{35.02}_{\hspace{0.4em}0.34\uparrow}$ & $\textbf{46.80}_{13.60\uparrow}$ & 33.20 & $\underline{40.80}_{\hspace{0.4em}7.60\uparrow}$ & $\textbf{59.60}_{11.20\uparrow}$ & 48.40 & $\underline{55.20}_{\hspace{0.4em}6.80\uparrow}$ & $\textbf{48.00}_{\hspace{0.4em}9.20\uparrow}$ & 38.80 & $\underline{43.60}_{\hspace{0.4em}4.80\uparrow}$ \\ 

 ID & $\textbf{58.17}_{16.40\uparrow}$ & 41.77 & $\underline{51.83}_{10.06\uparrow}$ & $\textbf{67.74}_{38.39\uparrow}$ & 29.35 & $\underline{50.11}_{20.76\uparrow}$ & $\textbf{44.83}_{13.80\uparrow}$ & 31.03 & $\underline{42.07}_{11.04\uparrow}$ & $\textbf{55.21}_{13.78\uparrow}$ & 41.43 & $\underline{49.38}_{\hspace{0.4em}7.95\uparrow}$ \\ 

 OOD & $\textbf{54.48}_{14.01\uparrow}$ & 40.47 & $\underline{47.81}_{\hspace{0.4em}7.34\uparrow}$ & $\textbf{53.07}_{20.09\uparrow}$ & 32.98 & $\underline{45.07}_{12.09\uparrow}$ & $\textbf{50.49}_{\hspace{0.4em}8.60\uparrow}$ & 41.89 & $\underline{44.51}_{\hspace{0.4em}2.62\uparrow}$ & $\textbf{52.42}_{11.49\uparrow}$ & 40.93 & $\underline{48.88}_{\hspace{0.4em}7.95\uparrow}$ \\

\bottomrule 
\end{tabular}
}
\setlength{\belowcaptionskip}{-3mm}
\caption{
The performance of the four base LLMs after fine-tuning under harmonious~(HAR), incompatible~(INC), and self-aligning~(SELF) settings. 
For each domain and base model, models fine-tuned on the harmonious dataset and on the self-aligning dataset achieve superior performance compared to those fine-tuned on the incompatible dataset, across all scenarios including homogeneous (HOMO), in-domain (ID), and out-of-domain (OOD) evaluations.
The best/second-best performance for each domain and base model in each evaluation is highlighted in bold/underline.
The arrows indicate the differences compared to the incompatible setting.
}
\label{table:exp1-1}
\end{table*}

In our first group of experiments, we would like to examine how the additional world knowledge in the IFT corpus affects LLMs.
To this end, we conduct experiments under three settings including harmonious, incompatible, and self-aligning.
By observing the performance discrepancies among these settings, we analyze the effects of injecting world knowledge into the IFT.

\hypertarget{finding1}{\paragraph{Finding 1.}{} \textit{When encompassing correct world knowledge, IFT data congruent with model parameter knowledge can lead to superior IFT outcomes.}}

\paragraph{}

To show this, we compare the experimental results under two settings: harmonious and incompatible.
Note that the datasets of both two settings contain correct world knowledge, meaning each query-response pair aligns with correct knowledge.  
Therefore, the core distinction between these two settings lies in whether the entailed world knowledge is consistent with the parameter knowledge of the LLM.
Specifically, training in the harmonious setting only requires learning behavioral norms without the need for learning any additional knowledge, while training in the incompatible setting requires learning both.

Table~\ref{table:exp1-1} reveals that models fine-tuned under the harmonious setting outperform those fine-tuned under the incompatible setting across homogeneous, in-domain, and out-of-domain evaluations\footnote{Regarding the evaluations of HOMO, ID, and OOD: The homogeneous, in-domain, and out-of-domain evaluations for each domain cover different subsets and the level of difficulty varies across these subsets. Therefore, the absolute performance differences between HOMO, ID, and OOD are not directly comparable.}.
Specifically, the harmonious setting yields mean performance gains of 11.27\%, 14.58\%, and 14.57\% over the incompatible setting for homogeneous, in-domain, and out-of-domain tests, respectively.
The results indicate that utilizing IFT data, which is consistent with model parameter knowledge and does not inject any additional domain-specific world knowledge, yields superior fine-tuned models. 
This conclusion holds true across homogeneous, in-domain, and out-of-domain evaluations.

\hypertarget{finding2}{\paragraph{Finding 2.}{} \textit{Using IFT data that aligns with model parameter knowledge yet is erroneous yields better performance than employing those that are correct but incongruent with model parameter knowledge.}}

\paragraph{}

To further investigate the impact of learning domain-specific world knowledge on IFT, we conduct a more direct comparative experiment between the self-aligning and incompatible settings.
For each domain and pre-trained LLM, the two settings' datasets use identical queries that the model cannot answer correctly under in-context learning and have different responses: the incompatible dataset's responses are correct, reflecting world knowledge, while the self-aligning dataset's responses represent model parameter knowledge, are incorrect.

Table~\ref{table:exp1-1} compares results under the self-aligning and incompatible settings. 
Surprisingly, despite the self-aligning dataset containing only incorrect answers, models fine-tuned on it significantly outperform those using the incompatible dataset, which requires learning inconsistent world knowledge.
The performance difference is notable, with the former achieving an average increase of 5.25\%, 9.78\%, and 6.97\% in homogeneous, in-domain, and out-of-domain evaluations, respectively.

This finding emphatically indicates that injecting additional domain knowledge through IFT also fails to bring effective improvements to LLMs even in homogeneous and in-domain evaluations.
Conversely, maintaining consistency with model parameter knowledge, that is, without injecting any additional world knowledge through the self-aligning setting, can yield superior results. 
Moreover, this advantage holds true across all evaluations, as well as different model sizes and architectures.

The above results demonstrate a significant decline of performance in models fine-tuned using data that contain correct world knowledge but conflict with model parameter knowledge, compared to using consistent IFT data aligned with parameter knowledge.
This suggests that introducing additional world knowledge through IFT, in cases where it is inconsistent with the parameter knowledge, may not yield the benefits we anticipate. 
Therefore, the core role of IFT may lie in facilitating the transfer of behavioral norms, rather than injecting additional domain-specific world knowledge. 
To further validate this conclusion, in the next section, we design a novel method to decouple the conflict knowledge contained in IFT data and present further analysis of this issue.

\section{Exp-\textup{II}: IFT with Contextualized Knowledge}
\label{sec:Contextualized IFT}

In this section, we introduce a new analysis method called contextualized knowledge decoupling to further investigate the impact of inconsistent knowledge during IFT. 
This approach involves explicitly providing relevant world knowledge needed to answer a query within the context of the query itself.
Under this paradigm, the model no longer needs to learn knowledge during IFT, but only needs to use the knowledge in the context to answer in the expected behavioral norm.
This method helps prevent the model from learning additional world knowledge during IFT, which is inconsistent with parameter knowledge, thus separating knowledge injection from behavioral norm transfer.

To this end, we start with the data of the incompatible setting to construct a dataset with contextualized knowledge.
Specifically, given an instruction-answer pair in the incompatible group, we employ GPT-3.5\footnote{\href{https://openai.com/}{https://openai.com/}} to generate the world knowledge that is required to answer the instruction. 
The knowledge is then concatenated with the original instruction, as well as the answer, to construct an augmented pair.
Finally, we use the constructed data to fine-tune LLMs and compare them with the models fine-tuned with vanilla IFT.

\begin{table}[!tp]
\resizebox{\columnwidth}{!}{
\begin{tabular}{llcccc}
\toprule

\textbf{Model}
& \textbf{IFT}
& \textbf{HOMO} 
& \textbf{ID}  
& \textbf{OOD}
& \textbf{Overall} \\

\midrule

\multirow{2}{*}{LLaMA-2-7B} 
 & Vanilla & 31.11 & 33.77 & 32.56 & 32.48 \\ 
 & Contextualized & \textbf{37.62} & \textbf{43.70} & \textbf{40.61} & \textbf{40.65} \\ 
\midrule

\multirow{2}{*}{LLaMA-2-13B} 
 & Vanilla & 34.51 & 37.35 & 37.24 & 36.37 \\ 
 & Contextualized & \textbf{41.47} & \textbf{49.48} & \textbf{46.59} & \textbf{45.84} \\ 
\midrule

\multirow{2}{*}{Mistral-7B} 
 & Vanilla & \textbf{38.77} & 35.90 & 39.07 & 37.91 \\ 
 & Contextualized & 37.22 & \textbf{43.47} & \textbf{44.99} & \textbf{41.89} \\ 
\midrule

\multirow{2}{*}{Average} 
 & Vanilla & 34.80 & 35.67 & 36.29 & 35.59 \\ 
 & Contextual & \textbf{38.77} & \textbf{45.55} & \textbf{44.06} & \textbf{42.79} \\

\bottomrule
\end{tabular}
}
\caption{
The performance comparison between models fine-tuned with vanilla IFT and models fine-tuned with contextualized IFT respectively.
}

\label{table:exp2-context}
\end{table}

\hypertarget{finding3}{\paragraph{Finding 3.}{} \textit{Ensuring that the model does not learn world knowledge conflicting with parameter knowledge during IFT enhances the effectiveness of IFT.}}
\paragraph{}

Table~\ref{table:exp2-context} shows the results of contextualized knowledge decoupling in three base models~\footnote{In this experimental setting, "vanilla" is equivalent to "incompatible" group.}.
From the table, it can be observed that fine-tuning the model with data using explicit contextualized knowledge significantly mitigates the adverse effects caused by inconsistencies between parameter knowledge and world knowledge in IFT data.
Compared to vanilla IFT using incompatible data, our method achieves an average improvement of 8.16\% on LLaMA-2-7B, 9.48\% on LLaMA-2-13B, and 3.98\% on Mistral-7B.
Except in rare cases, IFT using data with contextualized knowledge can significantly improve the effect of IFT in the incompatible setting across homogeneous, in-domain, and out-of-domain evaluations.

The results indicate that we should not force the model to learn additional inconsistent knowledge when the world knowledge in IFT data is inconsistent with model parameter knowledge.
Instead, by decoupling knowledge learning and the transfer of behavioral norms, the problems caused by the above knowledge conflicts can be effectively alleviated.
This further verifies our observation in Section~\ref{exp1} that additional world knowledge injection may be ineffective or even harmful for IFT.
During IFT, the model should focus on transferring the behavioral norms relying on the existing parameter knowledge and the regularity in the IFT data, rather than learning additional world knowledge.
Therefore, for \hyperlink{RQ1}{RQ1}, we conclude:

\paragraph{Conclusion 1.}{} \textit{For IFT, there is little, if not even causing additional damage, benefits from the learning of world knowledge incongruent with parameter knowledge.}
\paragraph{}

\hypertarget{exp3}{\section{Exp-\textup{III}: Is Consistency All You Need?}}
\label{exp3}

The above findings appear to suggest a conclusion: 
For better transfer of behavioral norms, we should employ IFT data that completely aligns with model parameter knowledge without any additional world knowledge.
To substantiate this hypothesis, we design a new set of experiments.

Specifically, by adjusting the proportion of samples derived from the incompatible and the self-aligning group, we aim to adjust the ratios of consistency between the world knowledge in the IFT data and model parameter knowledge, thereby observing the variations in IFT outcomes under different consistency ratios.

\hypertarget{finding4}{\paragraph{Finding 4.}{} \textit{Employing IFT data that is fully consistent with model parameter knowledge does not necessarily result in optimal performance.}}
\paragraph{}

\begin{figure}[!tp]
\centering
\includegraphics[width=\columnwidth]{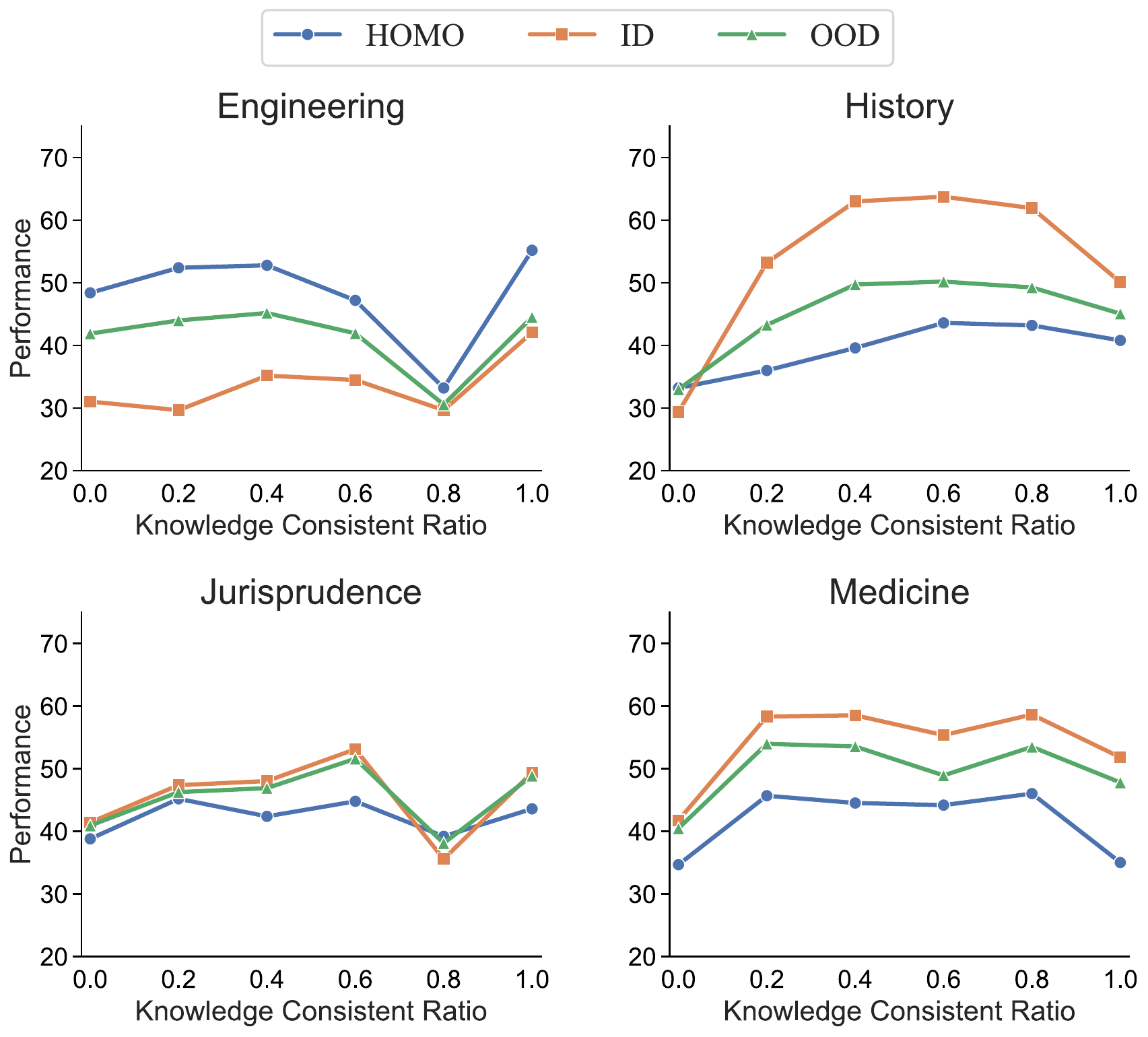}
\caption{The performance of Mistral-7B fine-tuned with instruction datasets of varying consistency ratios. 
Each dataset is composed of a mixture of incompatible and self-aligning data, and the consistency ratio represents the proportion of self-aligning samples. 
Note that a consistency ratio of 0 signifies that all data samples are incompatible, whereas a ratio of 1 indicates exclusively self-aligning data.
The results of other base models are presented in the Appendix~\ref{sec:appendix ratio-performance} due to page limitations.
}
\label{fig:ratio-perf}
\end{figure}

Figure~\ref{fig:ratio-perf} displays the results on three base models under different consistency ratios across four domains. 
From the figure, we can see that:
1) The performance of the incompatible group~(i.e., ratio=0) is indeed poor, which aligns with our previous conclusion;
2) Conversely, relying solely on IFT data that completely aligns with the model's parameter knowledge~(i.e., ratio=1) fails to ensure superior performance across a broad range of scenarios;
3) Optimal performance is most frequently achieved through a balanced integration of incompatible and self-aligning data. 
The ideal proportion of this combination varies across different base models and domains.

The observations suggest that while there is a significant impact of the consistency between the knowledge within the IFT data and the parameter knowledge of the original model on the performance of the fine-tuned model, this consistency is not the fundamental factor influencing IFT performance. 
Therefore, a deeper exploration into the mechanisms underlying knowledge consistency is essential to identify the true determinants of IFT's effectiveness.

\section{Exp-\textup{IV}: Rethinking Consistency: What Really Matters for IFT?}

To further explore the underlying mechanisms of IFT, we analyze the knowledge discrepancies of different base LLMs before and after IFT on various evaluation data to observe the extent of internal knowledge alternation triggered by IFT. 
Specifically, for each sample within the test data, we initially compute the Pearson correlation coefficient between the ranking of the original model's predictions on choices through in-context learning probing, and those provided by the fine-tuned model. 
Building upon this, we calculate the average Pearson correlation coefficient across each test set and subsequently compare it with the performance on the same test set of the fine-tuned model. 
Through this experimental analysis, we aim to observe how the knowledge changes induced by IFT affect the ultimate effect of IFT.

\begin{figure*}[t]

\centering
\includegraphics[width=0.8\textwidth]{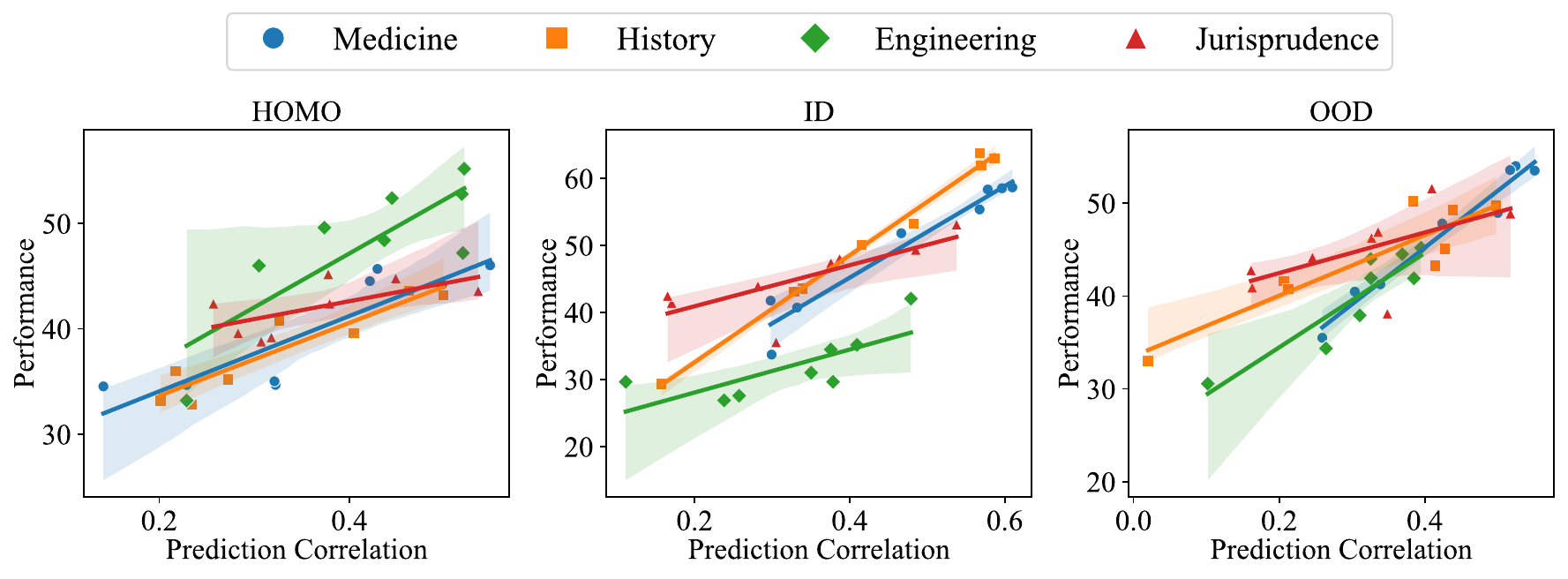}
\caption{
The regression analysis between the model performance after fine-tuning, and the knowledge consistency between base model and fine-tuned model. We show the results of Mistral-7B in three evaluations.
The grouped linear regression demonstrate the positive correlations between the model performance after IFT and model internal knowledge consistency before and after IFT.
Points in the same regression line indicate the results of the same base model fine-tuned with different IFT data of the same domain on the same test set (HOMO, ID, or OOD).
}
\label{fig:scatter}
\end{figure*}

To this end, we employ a partial correlation assessment~\cite{spearman1961proof} to analyze the aforementioned Pearson correlation and model performance.
Specifically, we utilize a total of 96 models, which include 72 models from \hyperlink{exp3}{Exp-\textup{III}}, in addition to models fine-tuned with consistency ratios of 0.05 and 0.1, conducting tests on homogeneous, in-domain, and out-of-domain data. 
To eliminate the potential influence of the base model's performance on the correlation analysis, ensuring that the results only reflect the differences brought about by varying IFT data, we treat the base model's performance as a control variable. 
By employing partial correlation assessment under the constraint of removing this control variable's influence, we analyze the correlation between the Pearson correlation coefficients of model predictions' rankings before and after IFT and their ultimate performance.
Please refer to the Appendix~\ref{sec:appendix analysis} for more details about our analysis.

\begin{table}[t]
\resizebox{\columnwidth}{!}{%
\begin{tabular}{lcccccc}
\toprule

\multirow{2}{*}{\textbf{Model}} 
& \multicolumn{2}{c}{\textbf{HOMO}}
& \multicolumn{2}{c}{\textbf{ID}} 
& \multicolumn{2}{c}{\textbf{OOD}} 
\\
\cmidrule(lr){2-3}
\cmidrule(lr){4-5}
\cmidrule(lr){6-7}

& \textbf{r}  & \textbf{p-value} & \textbf{r}  & \textbf{p-value} & \textbf{r}  & \textbf{p-value} \\
\midrule
Mistral-7B  &   0.78    &     0.00    &      0.81    &     0.00    &      0.82    &     0.00   \\
LLaMA-2-7B  &   0.27    &     0.14    &      0.19    &     0.30    &      0.21    &     0.24    \\
LLaMA-2-13B  &   0.56    &     0.00    &      0.78    &     0.00    &      0.87    &     0.00    \\
\midrule
All  &   0.43    &     0.00    &      0.57    &     0.00    &      0.48    &     0.00   \\ 
\bottomrule
\end{tabular}
}
\centering
\caption{
The Spearman partial correlation analysis between the model performance after fine-tuning, and the knowledge consistency between base model and fine-tuned model.
The analysis is controlled with the base model's performance on each test set.
r and p-value denote partial correlation coefficient and significance respectively. 
For LLaMA-2-13B and Mistral-7B, p-values significantly lower than 0.05 indicates a high level of confidence.
}
\label{tab:correlation}
\end{table}

\paragraph{Finding 5.}{} \textit{The consistency of internal knowledge within a model before and after IFT is the key factor affecting the performance of the fine-tuned model.
}
\paragraph{}

Figure~\ref{fig:scatter} presents the results of grouped regression analysis conducted on Mistral-7b. 
Table~\ref{tab:correlation} displays the correlation coefficients and significance for the partial correlation analysis across different base models and different evaluations. 
From the aforementioned figure and table, a significant trend can be observed, namely that the correlation of predictions made by models before and after IFT on a given evaluation has a substantial impact on the final performance of the fine-tuned models on that evaluation.
It is noteworthy that this phenomenon holds true across homogeneous, in-domain, and out-of-domain evaluations.
This implies that even for homogeneous and in-domain test sets, which are within the same field as the IFT data, the injection of additional world knowledge and the alteration of internal parameter knowledge through IFT do not contribute to enhancing the final performance of the models. 
Instead, maintaining consistency in the knowledge of models before and after IFT significantly positively influences the performance of the fine-tuned models.

\begin{table}[t]
\resizebox{\columnwidth}{!}{%
\begin{tabular}{lccc}
\toprule

\textbf{Model}
& \textbf{Best}
& \textbf{Self-aligning}
& \textbf{Incompatible}
\\
\midrule

Mistral-7B & 0.24 & 0.34 & 0.37 \\
LLaMA-2-7B & 0.16 & 0.93 & 0.51 \\
LLaMA-2-13B & 0.18 & 0.68 & 0.34 \\

\bottomrule

\end{tabular}
}
\centering
\caption{
KL divergence between the prediction distribution of fine-tuned model under zero-shot setting and the prediction distribution of base model using in-context learning probing.
"Best" denotes the model that exhibits the best average performance across three evaluations.
}
\label{tab:kl}
\end{table}

To further investigate whether this finding is the underlying mechanism responsible for \hyperlink{finding4}{Finding 4}, we explore the impact of different consistency data on the correlation of model knowledge before and after IFT in \hyperlink{exp3}{Exp-\textup{III}}. 
We analyze the probability distribution of model outputs under different IFT data conditions. 
Table~\ref{tab:kl} presents the related experimental results, which reveal that: 
IFT using data containing world knowledge completely inconsistent with the parameter knowledge evidently leads to a divergence in the internal knowledge of the fine-tuned model from that of the original model, thereby impairing the performance of the fine-tuned model. 
Furthermore, IFT using data containing world knowledge completely consistent with the parameter knowledge also may result in a divergence in the knowledge distribution between the original and fine-tuned models. 
Specifically, fine-tuning with only consistent IFT data may steer the model towards an sharp knowledge distribution, whereas the original model's parameter knowledge exists as a relatively smooth distribution. 
Conversely, using a middle setting that mixes incompatible and self-aligning data allows the optimization process to maintain model parameter knowledge unchanged while preserving the distribution's smoothness, thereby enabling the fine-tuned model outputs to more closely resemble those of the original model, ultimately yielding better performance.

The experimental results mentioned above demonstrate that the key to the performance of fine-tuned models lies in the consistency of model parameter knowledge before and after IFT.
In fact, the core mechanism underlying the superior performance observed in both \hyperlink{finding1}{Finding 1} and \hyperlink{finding2}{Finding 2} for the IFT data is attributed to the maintenance of consistency in model knowledge before and after IFT.
Therefore, for \hyperlink{RQ2}{RQ2} raised above, we conclude:

\paragraph{Conclusion 2.}{} \textit{The essence of an effective IFT lies in maintaining the consistency of model parameter knowledge before and after IFT.
}
\paragraph{}

\section{Conclusion and Discussion}
\label{sec:Conclusion}

Our experiments and conclusions indicate that the core function of IFT is not to learn domain-specific world knowledge. 
Instead, learning world knowledge that is inconsistent with model parameter knowledge actually undermines the performance of the model in all evaluations from homogeneous, in-domain to out-of-domain.
Furthermore, we discover that the consistency of model parameter knowledge before and after IFT (i.e., the knowledge probed through in-context learning before IFT and the knowledge exhibited under zero-shot setting after IFT) plays a crucial role in determining the ultimate performance of the fine-tuned model. 
These two findings unveil a fundamental mechanism of IFT, that is, IFT is not a supervised, domain-specific knowledge \textit{learning} process, but a process of \textit{self-aligning} instruction with the already existing parameter knowledge of LLMs. 
Therefore, the ultimate determinant of IFT effect is not the extent of domain knowledge injection, but rather whether the IFT process can facilitate more effective self-aligning, thereby enhancing the expression of the model's parameter knowledge under the zero-shot question-answering paradigm after IFT.

Our discovery not only provides guidance for future IFT data construction, model training, and model evaluation but also provide robust support for some very recent studies.
For instance, super-alignment~\cite{burns2023weaktostrong} aims to use a weak model to guide a strong model's alignment.
Our conclusion proves that it is entirely possible to use a weak model with less knowledge to guide a strong, more knowledgeable model for IFT.
Our conclusions elucidate the viable underlying factors for self-instruction-tuning~\cite{sun2024principle,guo2024human}, self-rewarding~\cite{yuan2024selfrewarding} and consistent alignment~\cite{wan2024mitigating} and provide a solid foundation for the future development of these studies.

\section*{Limitations}
\label{sec:Limitations}
In order to facilitate probing model parameter knowledge, we currently focus on multiple-choice questions. In the future, we plan to extend our framework to free-style generation. 
Besides, due to hardware limitations, the vast majority of experiments are conducted on models with about 10B parameters, we only explore 70B models in several experiments. Repeating our study on larger models in more domains will contribute to a deeper understanding of the IFT for larger models.

\section*{Acknowledgements}
We sincerely thank all anonymous reviewers for their insightful comments and valuable suggestions. This research work is supported by CAS Project for Young Scientists in Basic Research (Grant No.YSBR-040), the National Natural Science Foundation of China under Grants no. 62122077 and 62106251, and the Basic Research Program of ISCAS, Grant No. ISCAS-JCZD-202303.

\bibliography{rethinking}
\bibliographystyle{acl_natbib}

\onecolumn
\appendix

\section{Data Construction}
\label{sec:appendix Data Construction}

\subsection{Train}
As mentioned in Section~\ref{sec:framework}, we collect four multi-choice question datasets from different domains: medicine, history, engineering, and jurisprudence. 
The quantity and split of the dataset for each domain are detailed in Table~\ref{tab:source data}.

\begin{table}[!h]
\resizebox{0.35\columnwidth}{!}{%
\begin{tabular}{lccc}
\toprule

\textbf{Domain}
& \textbf{Dev}
& \textbf{Test}
& \textbf{Train}
\\
\midrule

Medicine & 10 & 1462 & 20000 \\
History & 10 & 250 & 8605 \\
Jurisprudence & 10 & 250 & 6510 \\
Engineering & 10 & 250 & 4805  \\

\bottomrule

\end{tabular}
}
\centering
\caption{The number of instances in the development, test, and train sets for each domain.}
\label{tab:source data}
\end{table}

For each item in each train set, we employ five demonstrations from the corresponding dev set for in-context learning to assess the parameter knowledge of each base model. The probing prompt is shown in Table~\ref{tab:infer format}.

\begin{table*}[!h]
    \centering
    \small
    \begin{tcolorbox}

    The following are multiple choice questions about <domain>. Please choose the correct answer. \\
    
    \textcolor{gray}{\textit{5-shot}} \\
    <question1> \\
    <question1 options> \\
    Answer:<question1 answer> \\
    
    <demo2> \\
    .\\
    .\\
    .\\
    <demo5> \\

    <question> \\
    <question options> \\
    Answer:\\

    \end{tcolorbox}
    \caption{Prompt design based on \citet{yang2023baichuan} for probing model parameter knowledge.}
    \label{tab:infer format}
\end{table*}

Because model's responses are influenced by the selection and order of demonstrations in in-context learning~\cite{gao2023ambiguity,min2022rethinking}, we regard responses with a confidence level exceeding 0.5 as reflective of the model's parameter knowledge to ensure the reliability of identification.
For every domain and base model combination, we establish train datasets on the three different settings as described in Section~~\ref{sec:framework}.
We report the amount of training data for each combination in Table~\ref{tab:train num}.
We employ the prompt in Table~\ref{tab:default IFT format} for constructing instruction-response pairs.

\begin{table}[!h]
\resizebox{0.7\columnwidth}{!}{%
\begin{tabular}{lcccc}
\toprule

\textbf{Model}
& \textbf{Engineering}
& \textbf{History}
& \textbf{Jurisprudence}
& \textbf{Medicine}
\\
\midrule

LLaMA-2-7B  & 738 & 996 & 1033 & 2507 \\
LLaMA-2-13B & 668 & 1236 & 676 & 1782\\
LLaMA-2-70B & 712 & 1072 & 1002 & 2518 \\
Mistral-7B  & 534 & 838 & 639 & 1400  \\

\bottomrule

\end{tabular}
}
\centering
\caption{The number of train sets for each domain and base model combination.}
\label{tab:train num}
\end{table}

\begin{table*}[!h]
    \centering
    \small
    \begin{tcolorbox}
    
    \textbf{\# Input} \\
    The following are multiple choice questions about {domain}. Please choose the correct answer. \\
    
    <question> \\
    <question options> \\
    
    \textbf{\# Output} \\
    <answer> \\
    Explanation: <explanation> \\
    
    \end{tcolorbox}
    \caption{Format of instruction-response pair for Vanilla IFT.}
    \label{tab:default IFT format}
\end{table*}

For <explanation> in the prompt, if the answer is consistent with the model's parameter knowledge but incorrect, we employ prompt in Table~\ref{tab:chat prompt} for prompting the base model to generate explanations for its choice; if the answer is correct, we directly use the explanation in MedMCQA for medicine and employ prompt in Table~\ref{tab:explanation prompt} for prompting GPT3.5 to generate explanations for the golden answer for the other three domains.

\begin{table*}[!h]
    \centering
    \small
    \begin{tcolorbox}
    
    \textbf{\# Instruction} \\

    Below is a list of conversations between a human and an AI assistant (you). \\
    Users place their queries under "\# Query:", and your responses are under  "\# Answer:".\\
    You are a helpful, respectful, and honest assistant.
    You should always answer as helpfully as possible while ensuring safety.\\
    Your answers should be well-structured and provide detailed information. They should also have an engaging tone.\\
    Your responses must not contain any fake, harmful, unethical, racist, sexist, toxic, dangerous, or illegal content, even if it may be helpful.\\
    Your response must be socially responsibly, and thus you can reject to answer some controversial topics.\\
    
    \textbf{\# Query:}\\
    \textasciigrave \textasciigrave \textasciigrave
    Can you tell me some common types of renewable energy sources? 
    \textasciigrave \textasciigrave \textasciigrave
    
    \textbf{\# Answer:}\\
    \textasciigrave \textasciigrave \textasciigrave \\
    Absolutely, below are some of the most common types of renewable energy sources:\\
    
    1. Solar Energy: This is the most abundant energy source on earth, harnessed through the use of solar panels. These panels convert sunlight into electricity without any moving parts, noise, pollution, or damage to the environment.\\
    2. Wind Energy: Wind turbines convert the kinetic energy in the wind into mechanical power. This mechanical power can be used for specific tasks (such as pumping water) or converted into electricity to power homes, businesses, and schools.\\
    3. Hydropower: Generated by using electricity generators to capture the energy from falling or fast-moving water. This renewable source can come from various ways, including a large dam on a river, tidal or wave energy from the ocean, or using small scale turbines in streams.\\
    4. Geothermal Energy: This type of energy is generated from the heat deep within the Earth. This heat can be used directly for heating buildings or to generate electricity. It is continuously produced inside the Earth and is nearly as reliable as the tides.\\
    5. Biomass Energy: Biomass is organic material that comes from plants and animals, and it contains stored energy from the sun. This energy can be burned directly or converted into biofuel which can burn more efficiently. \\
    
    Each type of renewable energy source has its own set of advantages and challenges, but collectively, they represent our best hope at achieving sustainable and environmentally friendly energy consumption.\\
    \textasciigrave \textasciigrave \textasciigrave \\
    
    \textbf{\# Query:} \\
    \textasciigrave \textasciigrave \textasciigrave \\
    Below is a multiple-choice question and the answer. Please give the explanation.\\
    Question: <question>\\
    Choices: <question options>\\
    Answer: <answer>\\
    \textasciigrave \textasciigrave \textasciigrave\\
    
    \textbf{\# Answer:} \\
    
    \end{tcolorbox}
    \caption{Prompt design based on URIAL~\cite{lin2023unlocking} for employing base model to generate explanations for its predicted answer.}
    \label{tab:chat prompt}
\end{table*}

\begin{table*}[!h]
    \centering
    \small
    \begin{tcolorbox}
    
    The following is a multi choice question about <domain>.\\

    <question>\\
    <choice>\\
    
    The answer is "<answer>". Please explain why.\\
    
    \end{tcolorbox}
    \caption{Prompt design for employing GPT-3.5 to generate explanations for the golden answer.}
    \label{tab:explanation prompt}
\end{table*}

For stable and real IFT, we incorporate an equal proportion of general instruction data sampled from alpaca-gpt4-en~\cite{peng2023instruction} which is identical for each domain and base model combination. 
We use vicuna-v1.5~\cite{zheng2024judging} format to train models.

\subsection{Test}
To compare the performance across three settings for each domain and base model combination, we devise three evaluation types: homogeneous, in-domain, and out-of-domain tests. 
The first evaluation involves a holdout from the same multi-choice question set, while the subsequent two, in-domain and out-of-domain, are derived from MMLU's splits.
The specific details of the splits for each domain are detailed in Table~\ref{tab:mmlu-split}.

\begin{table*}[!h]
\centering
\small
\begin{tabular}{p{0.15\textwidth}p{0.75\textwidth}}
\toprule
\textbf{Domain} & \textbf{In-domain Subcategories} \\
\midrule
Engineering  & \parbox[t]{0.73\textwidth}{
["electrical\_engineering"]
}\\
\midrule
History & \parbox[t]{0.73\textwidth}{
["high\_school\_european\_history", "high\_school\_us\_history",\\ "high\_school\_world\_history", "prehistory"]
}\\
\midrule
Jurisprudence & \parbox[t]{0.73\textwidth}{
["econometrics", "high\_school\_geography", "high\_school\_government\_and\_politics", \\ "high\_school\_macroeconomics", "high\_school\_microeconomics",\\ "high\_school\_psychology", 
"human\_sexuality", "international\_law", \\ "jurisprudence", "professional\_law", "sociology",
"public\_relations", \\
"professional\_psychology",
"security\_studies", "us\_foreign\_policy"]
}\\
\midrule
Medicine & \parbox[t]{0.73\textwidth}{
["anatomy", "clinical\_knowledge", "college\_medicine", "human\_aging", \\
"medical\_genetics", "nutrition", "professional\_medicine", "virology"]
}\\
\bottomrule
\end{tabular}
\caption{
In-domain and out-of-domain split of MMLU subcategories for our four experimental domains.
The out-of-domain subcategories comprise the remaining subcategories not included in the in-domain classification.}
\label{tab:mmlu-split}
\end{table*}

\section{Details about Contextualized IFT}
In Section~\ref{sec:Contextualized IFT}, we use the prompt in Table~\ref{tab:Contextualized IFT} to employ GPT-3.5 to provide the knowledge required for the inconsistent IFT data.
For contextualized IFT, we employ the prompt in Table~\ref{tab:Contextualized IFT format} for building instruction-response pairs to train models. 
Note we do not provide the knowledge required for the instruction during the testing phase.

\begin{table*}[!h]
    \centering
    \small
    \begin{tcolorbox}

    Given a multi-choice question and the answer, please write a short piece of evidence to support it so that a layman who has read the evidence you give can answer the question correctly. \\
    If your response contains words "listed", "option" or "choice" like "among the listed/given options', you will be penalized. \\

    \textbf{Question:} \\
    <question> \\
    <question options> \\
    
    \textbf{Answer:} \\
    <answer> \\
    
    \textbf{Evidence:} \\

    \end{tcolorbox}
    \caption{Prompt design for employing GPT-3.5 to generate the knowledge required for the instruction.}
    \label{tab:Contextualized IFT}
\end{table*}

\begin{table*}[!h]
    \centering
    \small
    \begin{tcolorbox}
    
    \textbf{\# Input} \\
    The following are multiple choice questions about <domain>. Given the context. Please choose the correct answer. \\
    
    <context> \\
    <question> \\
    <question options> \\
    
    \textbf{\# Output} \\
    <answer> \\
    
    \end{tcolorbox}
    \caption{Format of instruction-response pair for our contextualized IFT.}
    \label{tab:Contextualized IFT format}
\end{table*}

\section{Performance of Models Fine-tuned in Different Consistency Ratios}
\label{sec:appendix ratio-performance}
In \hyperlink{exp3}{Exp-\textup{III}}, due to page limitations, we only report the performance of Mistral-7B fine-tuned with instruction datasets of varying consistency ratios. In Figure~\ref{fig:ratio-perf}, we supplement the results of the other two base LLMs: LLaMA-2-7B and LLaMA-2-13B.

\begin{figure*}[!h]
\centering
\centering
\begin{subfigure}[b]{0.48\textwidth}
    \centering
    \includegraphics[width=\textwidth]{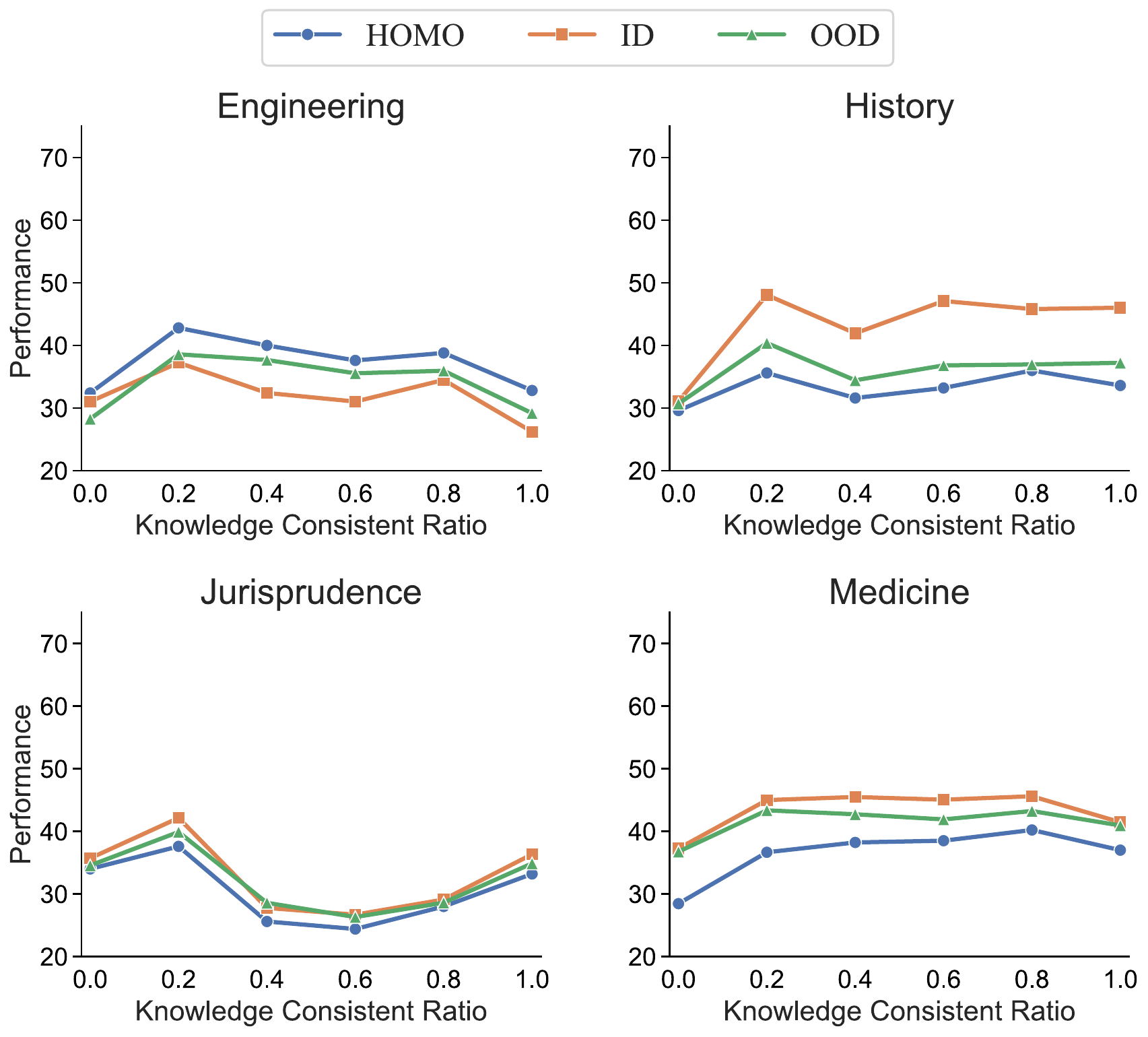} 
    \caption{LLaMA-2-7B}
    \label{fig:ratio performance LLaMA-2-7B}
\end{subfigure}
\hfill 
\begin{subfigure}[b]{0.48\textwidth}
\centering
\includegraphics[width=\textwidth]{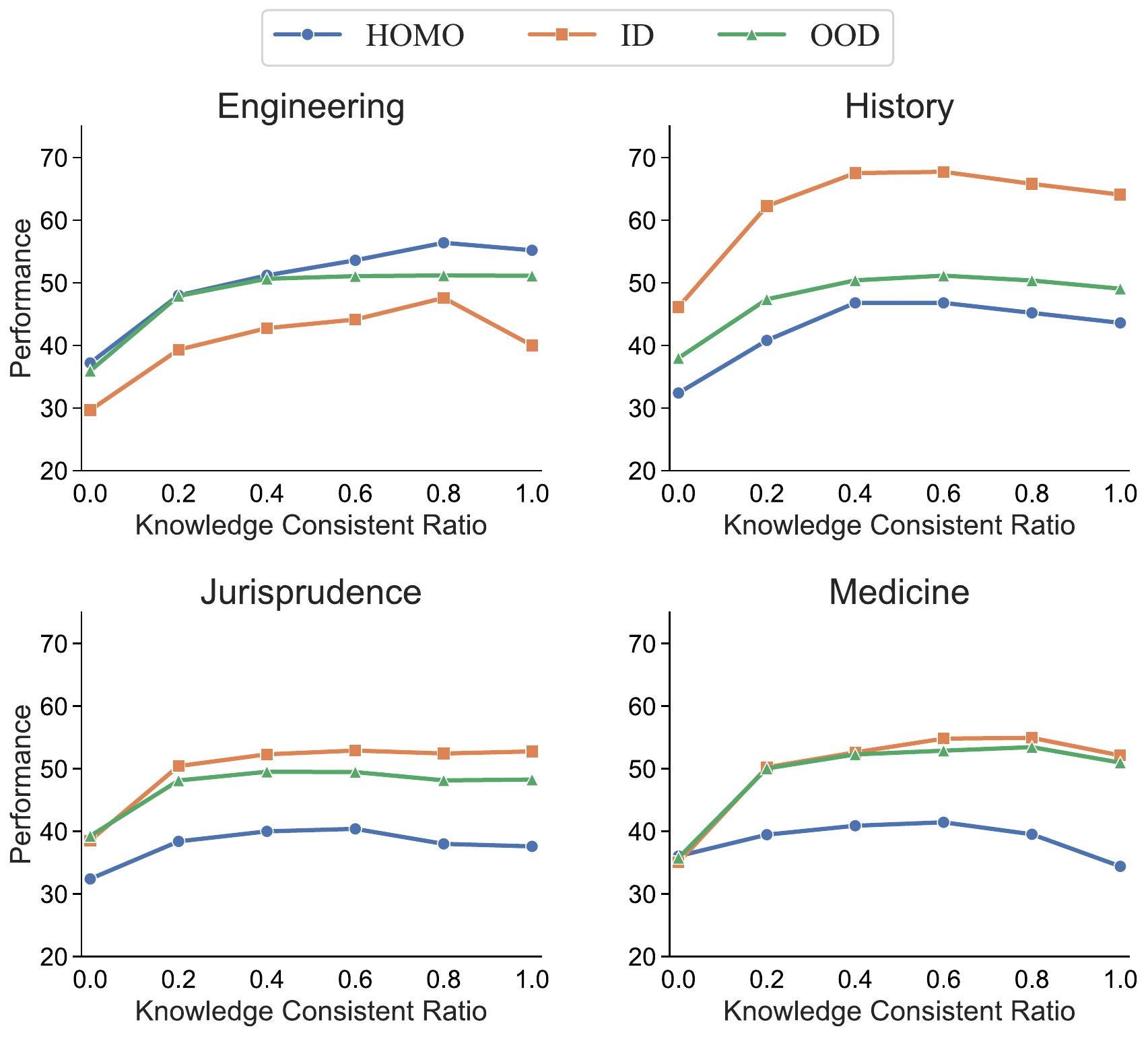} 
\caption{LLaMA-2-13B}
\label{fig:ratio performance LLaMA-2-13B}
\end{subfigure}
\caption{The performance of LLaMA-2-13B and LLaMA-2-7B fine-tuned with instruction datasets of varying consistency ratios. 
Each dataset is composed of a mixture of incompatible and self-aligning data, and the consistency ratio represents the proportion of self-aligning samples. 
}
\label{fig:ratio-perf}
\end{figure*}

\section{Details about Spearman Partial Correlation Analysis}
\label{sec:appendix analysis}
Partial correlation measures the degree of association between two variables (here, the model performance after IFT and the knowledge consistency between the base model and fine-tuned model) while controlling the effect of one or more additional variables (the accuracy on the same test set of the base model in our case).

To calculate the Spearman partial correlation between the model performance after IFT and the knowledge consistency between the base model and fine-tuned model, we first calculate the knowledge consistency, measured by the correlation coefficient of model prediction ranking before and after IFT using the following formula.
\begin{align}
pr=\frac{1}{n}\sum_{i=1}^{|D_{\text{test}}|}\text{pearson}(\text{sort}(m_{\text{base}},q_i,c_i),\text{sort}(m_{\text{tuned}}, q_i,c_i))    \nonumber
\end{align}

where $q_i$ and $c_i$ represent the i-th question and its corresponding set of candidate choices. The terms $m_{\text{base}}$ and $m_{\text{tuned}}$ denote the base model and its fine-tuned model, respectively. $D_{\text{test}}$ represents any test set like HOMO, ID, or OOD. $|D_{\text{test}}|$ denotes the size of the test set. The `sort` function ranks the candidate choices based on the predicted probability distribution from the input model. $pr$ reflects the average Pearson correlation coefficient between the rankings predicted by the base model and the fine-tuned model of each question. 

For each base model, its every fine-tuned model, and each test set, we calculate the corresponding above $pr$ and organize them along with the corresponding accuracy on the same test set of the base model and fine-tuned model to three lists, denoted as $pr_s$, $p_{\text{base}}$ and $p_{\text{tuned}}$. Then we compute the Spearman partial correlation. The formulas are:

\begin{align}
    e(pr_s)&=pr_s-\hat{pr_s}(p_{\text{base}}) \nonumber \\
    e(p_{\text{tuned}})&=p_{\text{tuned}}-\hat{p_{\text{tuned}}}(p_{\text{base}}) \nonumber \\ 
    r,p\text{-}value&=\text{spearman}(e(pr_s),e(p_{\text{tuned}})) \nonumber
\end{align}

Here, $\hat{pr_s}(p_{\text{base}})$ and $\hat{p_{\text{tuned}}}(p_{\text{base}})$ present the estimated value of $pr_s$ and $p_{\text{tuned}}$ predicted based on $p_\text{base}$, respectively. The residuals of $pr_s$ and $p_{\text{tuned}}$ relative to $p_{\text{base}}$ are denoted as $e(pr_s)$ and $e(p_{\text{tuned}})$, respectively. 
We finally calculate Spearman's correlation coefficient and the corresponding p-values between the residuals, denoted as $r$ and $p\text{-}value$. 

In this way, we can find the direct relationship between the model performance after IFT, and the knowledge consistency between the base model and fine-tuned model by removing the influence of base model performance that affects them both.

\end{document}